\documentclass[letterpaper]{article}
\usepackage{aaai2027}
\usepackage[hyphens]{url}
\usepackage{graphicx}
\graphicspath{{figures_libero_object/}{./}}
\usepackage{natbib}
\usepackage{caption}
\usepackage{amsmath,amssymb}
\usepackage{booktabs}
\usepackage{array}
\usepackage{colortbl}
\usepackage{xcolor}
\usepackage{amsmath, amssymb}
\usepackage{dsfont}

\definecolor{goodgreen}{RGB}{34,139,34}
\definecolor{headerbg}{RGB}{240,244,248}
\definecolor{oursbg}{RGB}{234,247,236}
\definecolor{bestbg}{RGB}{235,243,252}
\definecolor{goodgreen}{RGB}{34,139,34}
\definecolor{ablationbg}{RGB}{244,239,252}
\definecolor{badred}{RGB}{192,57,43}

\newcommand{\pionehalf}{\ensuremath{\pi_{0.5}}}
\newcommand{\method}[1]{\textsc{#1}}
\title{VLA-Corrector: Stage-Aware Observable State Understanding for Prompt-Based Closed-Loop Recovery of Vision-Language-Action Policies}

\author{
  Chang Song,\textsuperscript{\rm 1}
  Bin Qian,\textsuperscript{\rm 2}
  Yan Feng,\textsuperscript{\rm 3}
  Zhijie Song\textsuperscript{\rm 4}
}

\affiliations{
  \textsuperscript{\rm 1}Harbin Institute of Technology \\
  \textsuperscript{\rm 2}Tsinghua University \\
  \textsuperscript{\rm 3}Meituan \\
  \textsuperscript{\rm 4}Enacta AI \\
  \texttt{songchang@example.com},
  \texttt{qianbin@example.com},
  \texttt{fengyan@example.com},
  \texttt{zhijie.song@enacta.ai}
}
\affiliations{}

\begin{document}
\maketitle

\begin{abstract}
Long-horizon robot manipulation with Vision-Language-Action (VLA) policies remains vulnerable to execution-time deviations, as final task success provides little information for diagnosing and correcting failures caused by action noise, object displacement, or goal misalignment. We introduce a stage-aware failure verification and Prompt Recovery framework that enables closed-loop correction of a fixed VLA policy without parameter updates or privileged simulator states. The framework introduces an observable-history-based Learned Verifier that jointly estimates manipulation progress and execution risk by temporally modeling multi-view visual observations, proprioceptive states, and executed actions. To provide interpretable task understanding, we represent manipulation execution through semantic progress stages, including approach, alignment, grasp, transport, and placement, and identify stage-specific failure patterns. Upon detecting abnormal execution, the framework preserves the original instruction and generates a stage-conditioned recovery prompt, allowing the same frozen VLA policy to produce corrective actions. Extensive multi-round evaluations on LIBERO and LIBERO Plus demonstrate that the proposed approach substantially improves closed-loop reliability under diverse perturbations. Without access to privileged object or goal coordinates, the Learned Verifier achieves recovery performance close to that of the privileged rule-based verifier in the evaluated settings. These results show that observable visual-proprioceptive-action history is sufficient to infer latent task states and enable practical failure recovery for existing VLA policies.
\end{abstract}

\enlargethispage{3pt}

\section{Introduction}
\label{sec:introduction}

VLAs unify vision, language, and control to enable generalizable robotic manipulation via multimodal pretraining and in-flight replanning~\citep{brohan2023rt2,kim2024openvla,black2024pi0}. Yet, despite generating instruction-compliant actions, they lack mechanisms to verify execution fidelity, knowing \textit{what} to do but not \textit{whether} they are doing it correctly.

This limitation exposes a core gap between action generation and reliable embodied intelligence. 
In real-world manipulation, failures typically stem not from task misunderstanding, but from accumulated execution deviations during interaction. 
Minor action noise, object displacement, grasp imperfections, or goal-region shifts can progressively derail an otherwise valid trajectory. 
For long-horizon tasks, policies continuing unaware from such states often lack sufficient horizon for autonomous recovery. 
Biological agents mitigate this via continuous self-monitoring—evaluating not just \textit{what} to do, but \textit{whether} execution is proceeding correctly—highlighting execution-state awareness as a critical pillar for robust embodied agents.

In this work, we study execution-state understanding as a missing capability in current Vision-Language-Action systems. We investigate a fundamental question:

\begin{quote}
Can a robot agent infer its own execution state from observable interaction history and use this understanding to improve closed-loop reliability without modifying the underlying policy or accessing privileged environmental information?
\end{quote}

To address this, we introduce a lightweight framework atop a frozen VLA policy. A key verifier infers execution states from temporal histories of visual (agent/wrist), proprioceptive, and action data—bypassing privileged state access—to assess trajectory validity and risk. This design decouples policy competence from execution reliability: the VLA generates actions while the verifier handles monitoring and recovery.

Based on the inferred execution state, we further propose Prompt Recovery, a lightweight closed-loop correction mechanism that enables a frozen VLA policy to recover from execution deviations. When an anomaly is detected, the original task instruction is preserved and augmented with a stage-aware recovery prompt. The same VLA policy then generates corrective actions under the updated execution context. Unlike approaches that require policy finetuning, additional action heads, or direct trajectory modification, Prompt Recovery treats the VLA model as a reusable controller and introduces a simple interface between execution monitoring and action generation. This enables existing VLA policies to acquire recovery capabilities without changing their parameters.

To systematically evaluate execution-state understanding and recovery, we establish a controlled perturbation protocol on LIBERO and LIBERO Plus~\citep{liu2023libero}. The evaluation introduces reproducible execution deviations, including action noise, target-object shifts, and goal-region shifts, allowing us to analyze whether an agent can recognize execution risk and recover from it. We compare three matched settings: No Assistance, where the original VLA policy operates without monitoring; Rule + Prompt, where the privileged rule-based verifier provides the recovery interface using privileged state information; and Learned Binary + Prompt, where the Learned Verifier provides the same recovery interface without privileged information.

Across simulation benchmarks, Prompt Recovery improves standard LIBERO success
from $70.9\%$ to $82.6\%$, while the Learned Verifier
achieves $80.2\%$ without privileged state access. On LIBERO Plus, Rule +
Prompt and Learned Binary + Prompt reach $79.7\%$ and $79.0\%$ at difficulty
4, and $73.6\%$ and $72.6\%$ at difficulty 5, respectively. Under within-task
target-object shifts in the real-world experiments, our method
achieves an average success rate of $88.3\%$, compared with $82.5\%$ for the
frozen \texttt{pi05\_libero} baseline, demonstrating effective and transferable
closed-loop recovery.

Our contributions are summarized as follows:

\begin{enumerate}

\item We identify execution-state understanding as a fundamental missing capability in current Vision-Language-Action systems and formulate it as the bridge between action generation and reliable closed-loop embodied intelligence.

\item We introduce an observable-history-based execution-state verifier that learns to infer latent manipulation progress and execution risk from visual-proprioceptive-action trajectories without requiring privileged environmental states.

\item We propose Prompt Recovery, a lightweight policy-agnostic correction mechanism that enables frozen VLA policies to recover from execution deviations through stage-aware instruction augmentation without policy finetuning or action modification.

\item We establish a systematic perturbation-based evaluation protocol on LIBERO and LIBERO Plus to study execution monitoring and recovery capability of VLA policies under realistic interaction disturbances.

\end{enumerate}

\section{Related Work}
\label{sec:related}

\paragraph{Vision-Language-Action policies.}
Modern VLA systems map visual observations and language instructions to low-level control commands~\citep{brohan2022rt1,brohan2023rt2,driess2023palme,padalkar2023openxembodiment,kim2024openvla,black2024pi0,octo2024}, typically via action chunking or receding-horizon execution. Earlier visuomotor policies established scalable end-to-end control and offline manipulation learning~\citep{levine2016endtoend,kalashnikov2018qtopt,mandlekar2021robomimic}, building on established learning-from-demonstration and robot-learning formulations~\citep{schaal1999imitation,argall2009survey,kober2013reinforcement,billard2008robot,ravichandar2020survey}. Work on one-shot imitation, grasp synthesis, and self-supervised interaction further developed transferable visual manipulation primitives~\citep{duan2017oneshot,finn2017one,mahler2017dexnet,zeng2018pushing,bohg2014grasp}. Complementary generalist manipulation systems use multimodal prompting, affordance representations, or large-scale demonstrations to transfer across tasks~\citep{jang2022bcz,bousmalis2023robocat,wang2023vima,shridhar2023peract,mandlekar2024mimicgen,mees2022calvin}. While action chunking enhances temporal coherence and reduces policy query rates~\citep{zhao2023act,chi2023diffusion}, it introduces a re-planning latency during which outdated actions may persist despite environmental or kinematic changes. In this work, we freeze the base VLA policy and investigate how this latency can be monitored and mitigated using only observable execution histories.

\paragraph{Failure detection and execution monitoring.}
Execution-aware manipulation systems draw on geometric action representations and robust visuomotor control~\citep{pinto2017robust,zeng2021transporter,shridhar2022cliport}. Related work on temporal and demonstration-based visuomotor representation learning shows that visual context can support progress-sensitive manipulation decisions~\citep{sermanet2018timecontrastive,dasari2019robonet,xu2018neural,lynch2020play,nair2022r3m,goyal2023rvt}. In contrast, our monitor integrates temporal visual evidence with proprioceptive records and historical action sequences, emitting a binary execution-risk signal as the online trigger. We evaluate this trigger based on its impact on closed-loop task success rates, rather than relying solely on conventional detector metrics.

\paragraph{Robot recovery and language-guided recovery.}
Existing recovery strategies span low-level control, online replanning, expert demonstrations, and language-conditioned policies~\citep{ahn2022saycan,huang2022zeroshot,huang2022innermonologue,liang2023codeaspolicies,singh2023progprompt,huang2023voxposer}. Language-model agents have further demonstrated that explicit observations and self-generated feedback can guide action revision~\citep{yao2023react,shinn2023reflexion}. In contrast, the proposed Prompt Recovery framework retains the original task instruction while augmenting it with a state-aware recovery prompt. Crucially, recovery actions are synthesized by the same frozen \pionehalf{} policy, obviating the need for a separately trained recovery controller.

\paragraph{Benchmarks and scalable robot data.}
Generalization is commonly evaluated with multi-task robot-learning benchmarks such as Meta-World and RLBench~\citep{yu2020metaworld,james2020rlbench}, while large-scale teleoperation and cross-domain data collections expand the diversity of manipulations available for policy learning~\citep{mandlekar2018roboturk,ebert2021bridge,walke2023bridgedatav2,khazatsky2024droid}. LIBERO complements these settings by emphasizing knowledge transfer across sequential task suites~\citep{liu2023libero}. Our study uses this benchmark context to isolate the contribution of online verification and prompt-based recovery under controlled execution perturbations.

\paragraph{Privileged supervision and deployable observation.}
Simulation affords privileged state variables that are inherently amenable to progress verification. Domain and dynamics randomization aim to reduce the simulation-to-real gap for visuomotor policies~\citep{tobin2017domain,peng2018sim2real}, but do not by themselves yield a deployable execution monitor. We leverage simulator signals to implement the privileged rule-based verifier and generate corresponding offline teacher labels. In contrast, the online Learned Verifier operates exclusively on observable sequences of visual observations, proprioceptive states, and executed actions.

\section{Problem Formulation}
\label{sec:problem}

Let $q$ denote the original task instruction and let the current observation at time $t$ be
\begin{equation}
    o_t = (I_t^{a}, I_t^{w}, s_t),
\end{equation}
where $I_t^{a}$ and $I_t^{w}$ are agent-view and wrist-view RGB images, respectively, and $s_t\in\mathbb{R}^{8}$ is the observable proprioceptive state. Let $u_t\in\mathbb{R}^{7}$ be the action actually executed in the environment. The observable history available to the Learned Verifier is
\begin{equation}
    h_t = \{(I_i^{a},I_i^{w},s_i,u_i)\}_{i=0}^{t}.
\end{equation}
The history is temporally ordered: no observation or action from a future time step is used.

The fixed VLA policy maps the task instruction and the recent observation history to an action chunk:
\begin{equation}
    U_t = \pi_{\theta}(q,h_t),
\end{equation}
where $\theta$ is not modified by the proposed monitoring loop. Only a prefix of the chunk is executed before the next policy request, after which the policy is queried again using the updated observable history.

Let $f_t\in\{0,1\}$ denote the binary execution-risk label, where
$f_t=1$ indicates that a teacher-defined failure event occurs within the
prediction horizon. The Learned Verifier estimates
\begin{equation}
    p_t = \Pr(f_t=1\mid h_t), \qquad
    \hat f_t = \mathds{1}[p_t\geq\tau].
\end{equation}
The online gate applies consecutive-window confirmation and a cooldown period before accepting a trigger. If $k$ consecutive sampled windows exceed the threshold and the system is not in cooldown, a recovery prompt is issued. The threshold, confirmation length, cooldown, and sampling interval are selected on the validation split and fixed for the final evaluation.

The recovery prompt is
\begin{equation}
    q_t^{\mathrm{rec}} = q \oplus r_t,
\end{equation}
where $r_t$ is a stage-routed or generic recovery prompt fragment and $\oplus$ denotes instruction concatenation. The original task instruction remains present. The same policy generates recovery actions from $q_t^{\mathrm{rec}}$, after which the original task instruction is restored. We evaluate the final episode outcome, trigger quality, and action cost separately because a higher final success rate does not by itself establish higher binary detection accuracy.

\section{Method}
\label{sec:method}

\subsection{System Overview}
\label{sec:overview}
\begin{figure*}[t]
    \centering
    \includegraphics[width=\textwidth]{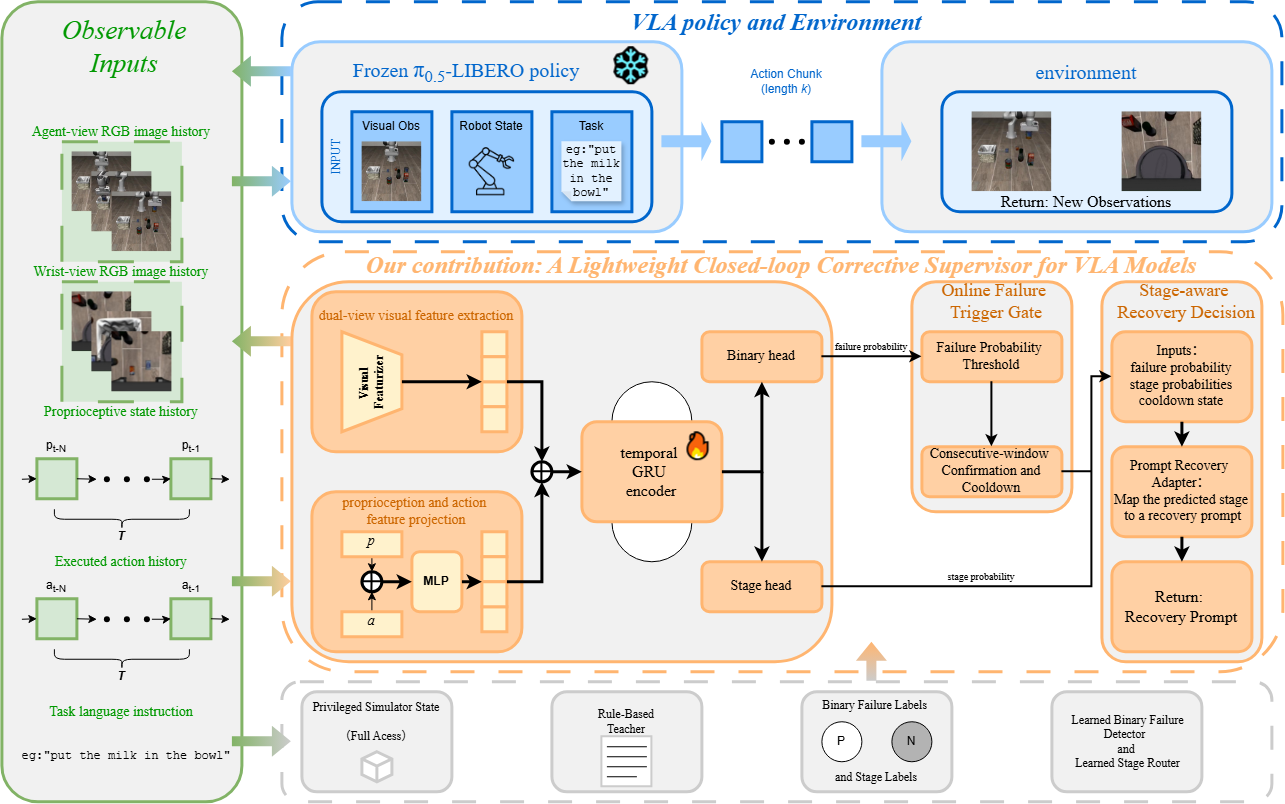}
    \caption{Overview of VLA-Corrector. The frozen VLA policy interacts with the environment using observable visual, proprioceptive, action, and language histories. A stage-aware verifier estimates execution risk and task stage, after which the trigger gate and recovery adapter generate a stage-specific recovery prompt. The privileged rule-based verifier provides supervision during verifier construction.}
    \label{fig:system-overview}
\end{figure*}

The overall architecture of \method{VLA-Corrector} is shown in
Fig.~\ref{fig:system-overview}.

The closed-loop correction process is summarized as
\begin{equation*}
\begin{aligned}
&\text{observation and instruction}
\rightarrow \pionehalf{}\text{ policy}
\rightarrow \text{action chunk}\\
&\rightarrow \text{risk verifier}
\rightarrow \text{trigger gate}
\rightarrow \text{Prompt Recovery}\\
&\rightarrow \text{task resumption}.
\end{aligned}
\end{equation*}
At each control cycle, the frozen policy outputs an action chunk, yet only a fixed-length prefix is executed before replanning. The verifier subsequently ingests the resulting observable history. If the gating mechanism suppresses an intervention, execution continues under the original task instruction. Otherwise, the pending action chunk is discarded, and a recovery action sequence is generated by querying the same policy with a recovery prompt. The original task resumes immediately upon completion of the recovery actions.
\subsection{Base \pionehalf{} Policy}
\label{sec:base-policy}

The base controller is the pretrained OpenPI \texttt{pi05\_libero} policy, accessed through the project inference interface with checkpoint \url{gs://openpi-assets/checkpoints/pi05_libero}. We keep this controller frozen throughout the recovery experiments: neither its parameters nor its visual encoder or action-generation head is updated. During evaluation, the client renders each LIBERO observation at $256\times256$ and resizes it to $224\times224$ before inference. Each server response is checked to ensure that it is a two-dimensional action array with action dimension 7. The policy returns an action chunk, of which the first five actions are executed before the next policy query (\texttt{replan\_steps=5}).

\subsection{Privileged Rule-Based Verifier}
\label{sec:rule}

The privileged rule-based verifier uses privileged state information to track the state machine
\begin{equation*}
\begin{aligned}
&\texttt{APPROACH}\rightarrow\texttt{ALIGN}\rightarrow\texttt{GRASP}\\
&\rightarrow\texttt{MOVE}\rightarrow\texttt{PLACE}\rightarrow\texttt{SUCCESS}.
\end{aligned}
\end{equation*}
The terminal \texttt{SUCCESS} state solely marks task completion; recovery is invoked only in prior stages. The privileged rule-based verifier utilizes simulator ground truth---encompassing object and region poses, end-effector kinematics, distances, gripper status, motion predicates, and task flags---to emit a structured recovery record (stage, context) upon rule-defined failures or timeouts. Since these signals are unavailable to the Learned Verifier during deployment, \method{Rule + Prompt} serves strictly as an efficacy upper bound, not a deployable monitor.

\subsection{Learned Verifier}
\label{sec:learned}

The Learned Verifier is the observable, deployable state-understanding module. At time $t$, it receives two RGB views, an 8-dimensional proprioceptive vector, and the 7-dimensional action actually executed in the environment. It forms an 8-step observable-history window sampled every two environment steps, and verifier images are stored at $112\times112$. Its visual branch is a parameter-free spatial-statistics featurizer. For each view, a $3\times4\times4$ adaptive-average-pooled RGB representation is concatenated with the RGB mean and standard deviation, yielding 54 features per view. The concatenated two-view representation is projected to 64 dimensions, and the 15-dimensional state-action vector is projected to 32 dimensions. A GRU with hidden dimension 64 integrates the temporal sequence. Two prediction heads then estimate execution risk and the current execution stage:
\begin{equation}
\begin{aligned}
    v_t &= \operatorname{Proj}_{64}(\operatorname{Visual}(I_{t}^{a},I_t^{w})),\\
    x_t &= \operatorname{Proj}_{32}([s_t;u_t]),\\
    z_t &= \operatorname{GRU}([v_t;x_t]),\\
    p_t&=\sigma(w_f^{\top}z_t+b_f),\\
    \ell_t^{\mathrm{stage}}&=W_{\mathrm{stage}}z_t+b_{\mathrm{stage}},\\
    \hat{c}_t&=\arg\max_{c\in\mathcal{C}}[\operatorname{Softmax}(\ell_t^{\mathrm{stage}})]_c.
\end{aligned}
\end{equation}
Here, $p_t$ and $\hat{c}_t$ denote the predicted execution-risk probability and execution stage, respectively. The Learned Verifier relies exclusively on observable inputs, excluding privileged simulator states (e.g., object poses, BDDL predicates, success flags), which are used only offline for supervision. Consequently, the learned representation acts as an observable proxy, where spatiotemporal visual dynamics, proprioception, and actions collectively indicate progression through stages (approaching, grasping, placing) or transitions into execution-risk states.

The stage prediction uses the five nonterminal classes $\mathcal{C}=\{\texttt{APPROACH},\texttt{ALIGN},\texttt{GRASP},\texttt{MOVE},\texttt{PLACE}\}$; the terminal \texttt{SUCCESS} state is handled by the environment outcome rather than by the learned stage head.

\paragraph{Teacher supervision and learned alignment.}

The teacher does not provide privileged coordinates as regression targets. Instead, it converts privileged task-progress state into operational labels that define the semantics learned by the verifier. Let $c_t$ denote the teacher-derived execution stage, and let $\mathcal{E}$ denote the set of teacher-defined failure-event steps. The stage head learns
$c_t\in\{\texttt{APPROACH},\texttt{ALIGN},\texttt{GRASP},\texttt{MOVE},\texttt{PLACE}\}$, while the binary head learns execution risk:
\begin{equation}
 f_t=
 \begin{cases}
 1, & \min_{e\in\mathcal{E}}|e-t|\leq 10,\\
 0, & \min_{e\in\mathcal{E}}|e-t|>20,
 \end{cases}
\end{equation}
Windows in the ambiguous intermediate region are excluded from binary training, aligning the Learned Verifier with stage semantics and execution risk—rather than object or goal coordinates. At deployment, the binary probability governs trigger-gating, while the stage estimate supplies contextual cues for recovery routing. Consequently, the stage head acts solely as a router, not a five-class risk detector.

\subsection{Online Trigger Gate and Recovery Routing}
\label{sec:gate}

Binary predictions are converted to interventions via thresholding ($\tau=0.40$), four-frame confirmation, and a 10-step cooldown. This pipeline suppresses transient false positives while preventing intervention redundancy. Upon triggering, the predicted stage $r_t$ routes context-specific prompts (e.g., approach, grasp); low-confidence estimates ($<0.9625$) default to a generic recovery prompt. All gate parameters are fixed post-validation, with verifier queries issued every two environment steps.

\subsection{Prompt Recovery}
\label{sec:prompt}

Both verifiers share one Prompt Recovery interface. Given $q$ (e.g., ``pick up the milk...''), the system forms $q\!\oplus\!r_t$ by appending ``Immediate recovery: '' and a stage directive derived from the verifier record. The frozen \pionehalf{} policy handles all executions. Restoring $q$ post-recovery, this setup isolates the verifier-trigger pathway as the sole variable; performance gaps thus reflect differences in detection timing, routing, and intervention. Low-confidence cases default to \texttt{UNKNOWN\_FAILURE}.

\section{Experimental Setup}
\label{sec:setup}

\subsection{Benchmark and Evaluation Protocol}

We evaluate the default OpenPI \texttt{pi05\_libero} policy under three benchmark settings: standard LIBERO, LIBERO Plus at difficulty 4, and LIBERO Plus at difficulty 5. Standard LIBERO comprises LIBERO-SPATIAL, LIBERO-GOAL, LIBERO-OBJECT, LIBERO-10, and LIBERO-90. LIBERO Plus adopts the corresponding Spatial, Goal, Object, and 10 suites at difficulty levels 4 and 5. The three operational configurations are summarized in Table~\ref{tab:operational-configurations}. All configurations share the same frozen VLA policy, while the assisted variants employ a unified Prompt Recovery interface.

\begin{table}[!ht]
\centering
\scriptsize
\setlength{\tabcolsep}{3pt}
\renewcommand{\arraystretch}{1.08}
\caption{Operational configurations.}
\label{tab:operational-configurations}

\begin{tabular}{@{}
>{\raggedright\arraybackslash}p{0.24\linewidth}
>{\raggedright\arraybackslash}p{0.29\linewidth}
>{\raggedright\arraybackslash}p{0.25\linewidth}
@{}}
\toprule
Configuration & Online verifier & Prompt Recovery \\
\midrule
No Assistance
& None; original frozen policy
& No \\

Rule + Prompt
& Privileged rule-based verifier
& Yes; shared interface \\

Learned Binary + Prompt
& Learned Verifier
& Yes; shared interface \\
\bottomrule
\end{tabular}
\end{table}

\subsection{Perturbation Purpose and Definition}

We evaluate controlled shifts via action noise and spatial perturbations. The clean condition has no perturbation; action noise adds zero-mean Gaussian noise ($\sigma=0.1$) to normalized actions, while target-object and goal-region shifts displace the corresponding XY coordinates by $0.1\,\mathrm{m}$. The combined condition applies all three perturbations; image noise is disabled (\texttt{action\_noise\_std}=0.1, \texttt{image\_noise\_std}=0.0, \texttt{perturb\_target\_xy}=\texttt{perturb\_goal\_xy}=0.1).

Methods share initial states, seeds, and perturbation protocols, with shuffling disabled. We use \texttt{replan\_steps}=5 and \texttt{recovery\_replan\_steps}=3, cap normal execution at 280 steps, and allow up to 100 actions per Prompt Recovery intervention.

\subsection{Metrics and Learned Verifier Configuration}

The primary metric is episode success rate (successful episodes divided by total attempts). For the Learned Verifier, we report Precision, Recall, Positive F1, AUROC, and AUPRC, with execution risk or a teacher warning as the positive class. Gate quality, recovery success, and action cost are reported separately to isolate the closed-loop components.

Rather than predicting final task success, the Learned Verifier estimates execution risk or a teacher warning from observable history. Its inputs, data split, and optimization settings are summarized in Table~\ref{tab:verifier-configuration}.

At deployment, the gate uses a risk threshold of 0.40, four consecutive positive windows, a 10-sample cooldown, and evaluation every two environment steps. Route selection uses a stage-confidence threshold of 0.9625; low-confidence cases are assigned to \texttt{UNKNOWN\_FAILURE}. The offline detector threshold (0.03294) is distinct from the online gate threshold.

\begin{table}[t]
\centering
\caption{Learned Verifier training and evaluation configuration.}
\label{tab:verifier-configuration}
\scriptsize
\begin{tabular}{p{0.43\linewidth}p{0.37\linewidth}}
\toprule
Parameter & Value \\
\midrule
Collected episodes & 120, episode-disjoint \\
Episode-disjoint split & 60\% train / 20\% validation / 20\% test \\
Sequence length & 8; single-frame ablation uses 1 \\
Prediction horizon & 10 environment steps \\
Negative exclusion & 20 environment steps \\
Training epochs & 30 \\
Batch size & 32 \\
Learning rate & $10^{-3}$ \\
Weight decay & $10^{-4}$ \\
Device & CPU \\
Input & Dual-view RGB + 8D proprioception + 7D action history \\
\bottomrule
\end{tabular}
\end{table}

\subsection{Real-World Evaluation}
\label{sec:real-world-evaluation}

To assess real-world transfer, we deploy the same frozen OpenPI
\texttt{pi05\_libero} policy and closed-loop correction framework on a
physical robot platform. All methods use the same robot, end-effector,
visual sensors, state-observation interface, control frequency, and
data-processing pipeline.

We evaluate three representative manipulation tasks:
\begin{enumerate}
    \item \textbf{Pick and Open:} The robot grasps a toy drawer and pulls it
    open.

    \item \textbf{Place:} The robot places a small wooden block into a
    designated compartment of a wooden four-compartment box.

    \item \textbf{Pour Water:} The robot grasps a plastic bottle and tilts it
    to pour the liquid into a container.
\end{enumerate}

\begin{figure*}[t]
\centering
\includegraphics[width=\textwidth]{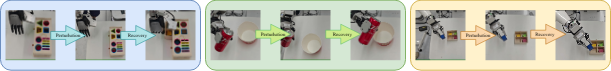}
\caption{Real-world manipulation tasks used for evaluation: Pick and Open,
Place, and Pour Water.}
\label{fig:real-world-tasks}
\end{figure*}

The three real-world task configurations are illustrated in
Fig.~\ref{fig:real-world-tasks}.

To evaluate recovery capability, we introduce a controlled spatial
perturbation during each task by deliberately changing the position of the
target object after execution has begun. The system is then evaluated
according to whether it can recover from the perturbation and successfully
complete the task.

Each task is evaluated over 40 independent trials. No real-world data are used for policy fine-tuning or verifier training.
No real-world data are used for policy finetuning or verifier retraining. We report the execution success rate, defined as the proportion of successful trials among all
evaluation trials.

\section{Results}
\label{sec:results}

\subsection{Overall Performance in Simulation}
\label{sec:overall-performance}

We compare three benchmark-level dimensions: standard LIBERO, LIBERO Plus at difficulty 4, and LIBERO Plus at difficulty 5. Standard LIBERO aggregates LIBERO-SPATIAL, LIBERO-GOAL, LIBERO-OBJECT, LIBERO-10, and LIBERO-90. The corresponding success rates are 70.9\%, 82.6\%, and 80.2\% for No Assistance, Rule + Prompt, and Learned Binary + Prompt, respectively.

For LIBERO Plus difficulty 4, the average success rates across the clean and perturbed conditions are 73.7\%, 79.7\%, and 79.0\%. For difficulty 5, aggregating the reported LIBERO-Object and LIBERO-10 conditions, the corresponding rates are 66.7\%, 73.6\%, and 72.6\%. Prompt Recovery consistently improves performance over No Assistance, while the Learned Verifier remains close to the privileged rule-based verifier.

\begin{figure}[t]
\centering
\includegraphics[width=\columnwidth]{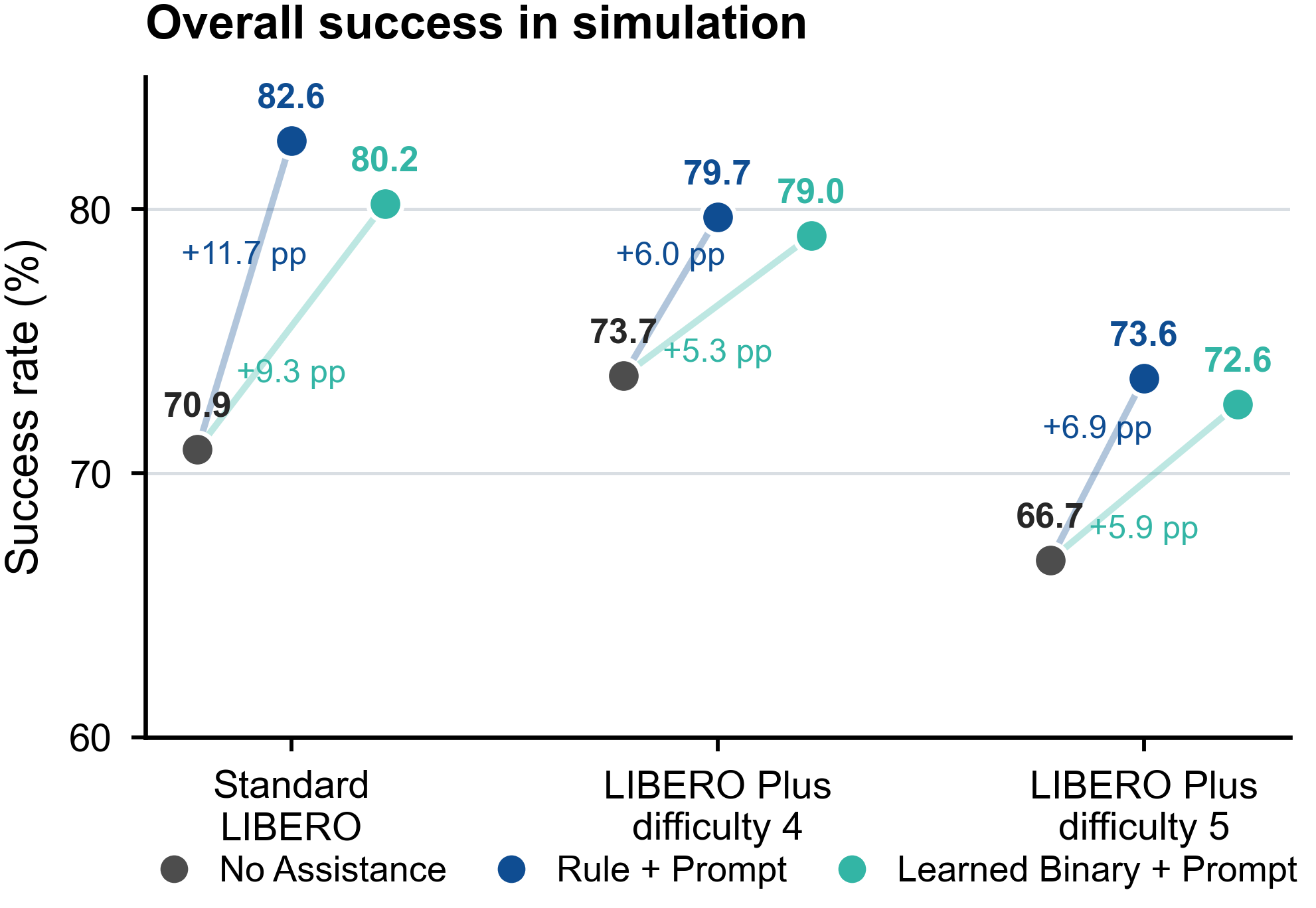}
\caption{Success rates in simulation across standard LIBERO and LIBERO Plus at difficulty levels 4 and 5.}
\label{fig:results-overall-performance}
\end{figure}

The corresponding comparisons are shown in
Fig.~\ref{fig:results-overall-performance}.

\subsection{Robustness to Environmental Perturbations}
\label{sec:perturbation-results}

Across clean execution, action noise, target-object shift, goal-region shift, and combined perturbations, both Prompt Recovery methods consistently outperform No Assistance. The performance gains are most pronounced under target-object and goal-region shifts, with the combined condition yielding success rates of 70.9\%, 82.6\%, and 80.2\% for No Assistance, Rule + Prompt, and Learned Binary + Prompt, respectively. The same ordering is maintained across the reported benchmark conditions, demonstrating that stage-aware Prompt Recovery improves robustness to diverse execution disturbances.

The complete results are reported in
Table~\ref{tab:perturbation-robustness}.

\begin{table*}[t]
\centering
\scriptsize
\setlength{\tabcolsep}{3pt}
\renewcommand{\arraystretch}{1.12}
\caption{Robustness to environmental perturbations on standard LIBERO.
Each entry reports mean success rate (\%) over 1000 evaluation episodes under
the specified condition. Arrows indicate absolute improvement over
No Assistance.}
\label{tab:perturbation-robustness}

\begin{tabular}{lcccccc}
\toprule
\rowcolor{headerbg}
\textbf{Method}
& \textbf{Clean}$\uparrow$
& \textbf{Action Noise}$\uparrow$
& \textbf{Target-Object Shift}$\uparrow$
& \textbf{Goal-Region Shift}$\uparrow$
& \cellcolor{bestbg}\textbf{Combined}$\uparrow$
& \textbf{Avg.}$\uparrow$ \\
\midrule

No Assistance
& 93.7
& 85.2
& 78.3
& 77.6
& \cellcolor{bestbg}70.7
& 81.14 \\

Rule + Prompt
& 94.2
& 89.4
& 86.2
& 88.2
& \cellcolor{bestbg}82.6\textcolor{goodgreen}{$\uparrow\,11.7$}
& 88.12\textcolor{goodgreen}{$\uparrow\,6.98$} \\

\rowcolor{oursbg}
\textbf{Learned Binary + Prompt}
& \textbf{94.1}
& \textbf{88.9}
& \textbf{83.7}
& \textbf{87.5}
& \cellcolor{bestbg}\textbf{80.2}\textcolor{goodgreen}{$\uparrow\,9.3$}
& \textbf{86.88}\textcolor{goodgreen}{$\uparrow\,5.74$} \\

\bottomrule
\end{tabular}
\end{table*}

\subsection{Failure Detection and Recovery Analysis}

The Learned Verifier achieves 93.1\% accuracy, 61.1\% precision, 94.9\% recall, 74.4\% Positive F1, AUROC of 0.902, and AUPRC of 0.7123. Its 0.457\,ms (p50) and 0.589\,ms (p95) inference latency enables efficient online monitoring without privileged state. Combined with the gating mechanism, it achieves 79.2\% trigger precision and supports safe recovery. Diagnostic logs show recovery success rates of 77.0\% for Rule + Prompt and 75.0\% for Learned Binary + Prompt, confirming the effectiveness of the shared Prompt Recovery interface.

\begin{figure}[t]
\centering
\includegraphics[width=\columnwidth]{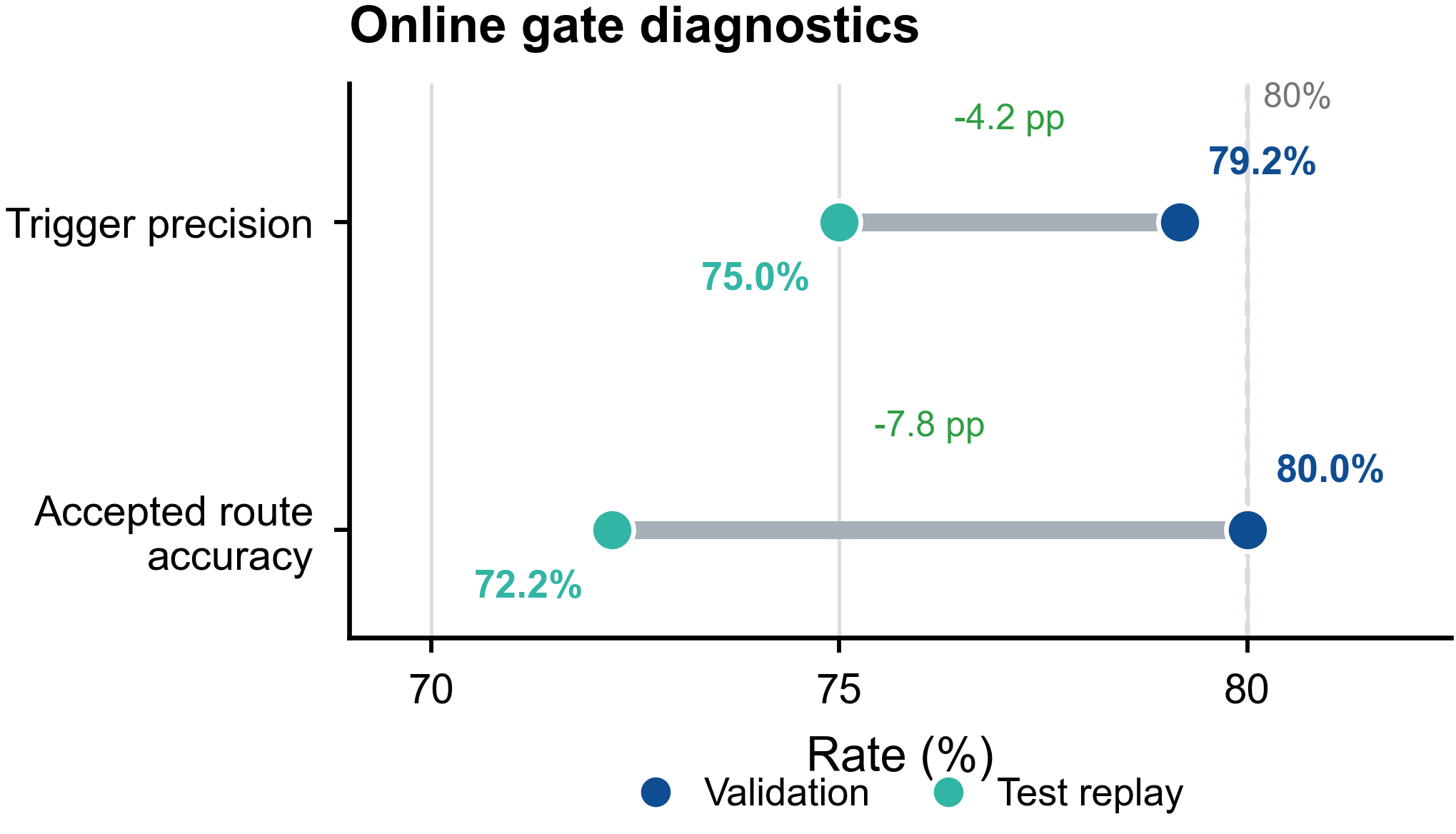}
\caption{Online gate diagnostics.}
\label{fig:results-detection-recovery}
\end{figure}

The corresponding validation and replay diagnostics are shown in
Fig.~\ref{fig:results-detection-recovery}.

\subsection{Ablation Studies}

The detailed ablation results are summarized in
Table~\ref{tab:verifier-ablation}.

The ablation study isolates the contributions of temporal history, camera observations, and state-action information while keeping the model structure and training protocol fixed. Temporal + Fusion provides the strongest execution-risk recall and Positive F1, demonstrating the value of combining temporal evidence with complementary visual and state-action inputs. Vision only is the strongest masked-input alternative and retains a favorable precision--recall trade-off, whereas Agent-view only is weaker than the full fusion configuration. State-action only performs worst overall, indicating that proprioceptive and action history alone are insufficient for reliable execution-state understanding. All configurations maintain sub-millisecond inference latency, supporting the feasibility of online monitoring around a frozen VLA policy.

\begin{table}[t]
\centering
\scriptsize
\setlength{\tabcolsep}{2.2pt}
\renewcommand{\arraystretch}{1.05}
\caption{Ablation study of the Learned Verifier.
Arrows indicate the absolute change in Positive F1 relative to Temporal + Fusion.}
\label{tab:verifier-ablation}

\begin{tabular}{@{}lccccc@{}}
\toprule
\rowcolor{headerbg}
\textbf{Configuration}
& \textbf{Acc.}$\uparrow$
& \textbf{Prec.}$\uparrow$
& \textbf{Recall}$\uparrow$
& \textbf{Pos. F1}$\uparrow$
& \textbf{p50/p95 (ms)}$\downarrow$ \\
\midrule

Single frame
& 86.95
& 48.87
& 75.49
& 59.33 \textcolor{badred}{$\downarrow\,15.02$}
& 0.214/0.376 \\

Agent-view only
& 89.73
& 52.33
& 80.92
& 63.56 \textcolor{badred}{$\downarrow\,10.79$}
& 0.420/0.503 \\

Vision only
& 92.66
& 58.00
& 82.24
& 68.03 \textcolor{badred}{$\downarrow\,6.32$}
& 0.371/0.477 \\

State-action only
& 80.75
& 47.67
& 75.88
& 58.55 \textcolor{badred}{$\downarrow\,15.80$}
& 0.382/0.455 \\

\midrule
\rowcolor{ablationbg}
\textbf{Temporal + Fusion}
& \textbf{93.09}
& \textbf{61.14}
& \textbf{94.85}
& \textbf{74.35}
& \textbf{0.457/0.589} \\

\bottomrule
\end{tabular}
\end{table}

\subsection{Real-World Evaluation Results}

\begin{figure}[t]
\centering
\includegraphics[width=\columnwidth]{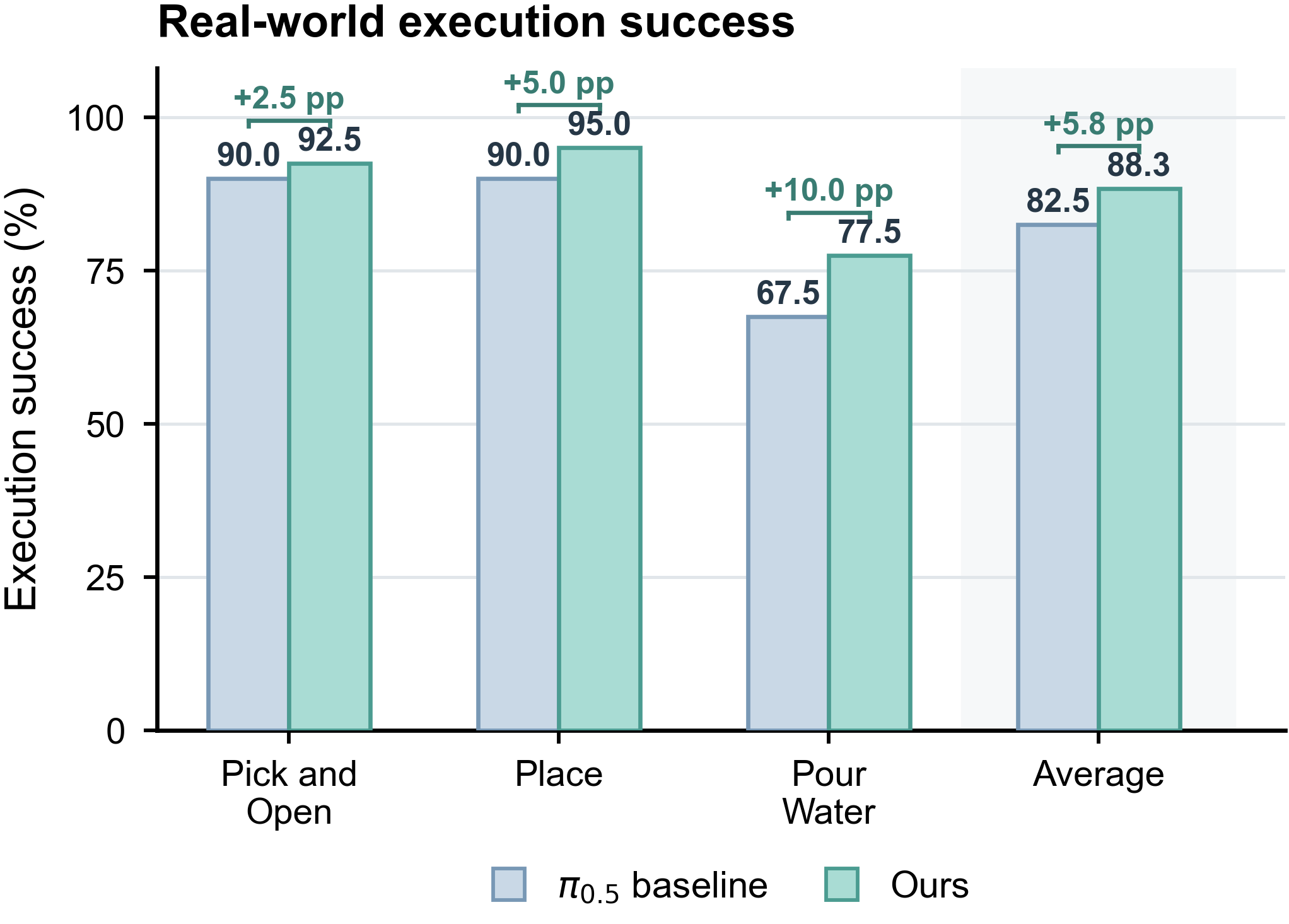}
\caption{Execution success rates of the frozen $\pi_{0.5}$ baseline and the proposed method across three real-world manipulation tasks. The annotations indicate absolute improvements in percentage points.}
\label{fig:real-world-success}
\end{figure}

Across the three real-world manipulation tasks, the proposed method achieves
an average execution success rate of $88.3\%$, compared with $82.5\%$ for the
frozen \texttt{pi05\_libero} baseline, corresponding to an absolute
improvement of $5.8$ percentage points. The task-level success rates are
reported in Fig.~\ref{fig:real-world-success}.

\section{Discussion}
\label{sec:discussion}

Our results demonstrate that effective failure recovery does not require privileged state information, such as object coordinates or BDDL predicates. By operating solely on observable history, the Learned Verifier achieves performance comparable to the privileged rule-based verifier (within 2.4 pp) while enabling a non-intrusive interface to a frozen VLA policy. This decoupling allows the recovery module to function as a lightweight monitor, separating state estimation from the control policy and avoiding the need for costly re-training.

The consistent gains across action noise, target-object shifts, and goal-region shifts (+11.7 pp and +9.3 pp over the No Assistance baseline) validate the robustness of this paradigm. The narrow performance gap between the Learned Verifier and the privileged rule-based verifier suggests that high-frequency execution-risk detection can be reliably delegated to data-driven modules. This work thus establishes a practical pathway for deploying closed-loop recovery mechanisms in real-world settings where ground-truth state information is unavailable.

\section{Conclusion}
\label{sec:conclusion}

We identify execution-time opacity in action-chunked VLA policies, whereby
plausible action sequences can enter execution-risk states under perturbations.
\method{VLA-Corrector} introduces a non-invasive detect-and-correct module
that monitors temporally ordered observation--proprioception--action histories
to estimate execution risk and trigger stage-aware Prompt Recovery, while
keeping the base policy frozen. These results show that observable visual-proprioceptive-action history can
provide useful estimates of execution stage and execution risk, enabling
practical recovery for existing VLA policies.

\bibliography{references}

\end{document}